\pdfoutput=1
\documentclass[11pt]{article}

\usepackage{acl}

\usepackage{times}
\usepackage{latexsym}

\usepackage[T1]{fontenc}
\usepackage[utf8]{inputenc}

\usepackage{microtype}

\usepackage{graphicx}
\usepackage{amsmath}
\usepackage{amsfonts}
\usepackage{listings}
\usepackage{multirow}
\usepackage{subfig}
\usepackage{hyperref}
\let\orgautoref\autoref
\renewcommand{\autoref}
{\def\equationautorefname{\emph{Eq. }\!\!}%
	\def\figureautorefname{\emph{Figure }\!\!}%
	\def\subfigureautorefname{\emph{Fig.}\!\!}%
	\def\algorithmautorefname{\emph{Algorithm}}%
	\def\sectionautorefname{\S\!\!}%
	\def\subsectionautorefname{\S\!\!}%
	\def\subsubsectionautorefname{\S\!\!}%
	\def\Itemautorefname{\emph{item}}%
	\def\tableautorefname{\emph{Table}}%
	\orgautoref}

\newcommand{\tabincell}[2]{\begin{tabular}{@{}#1@{}}#2\end{tabular}}
\title{Gated Mechanism Enhanced Multi-Task Learning for Dialog Routing}

\author{
    Ziming Huang$^1$\thanks{\text{}  This work was done when they were in IBM Research China, and Zhuoxuan Jiang is the corresponding author.}, Zhuoxuan Jiang$^1$\footnotemark[1], Ke Wang$^2$\footnotemark[1], Juntao Li$^3$, Shanshan Feng$^4$, Xian-Ling Mao$^5$ \\ $^1$Tencent, $^2$Alibaba DAMO Academy,  $^3$Soochow University, \\ $^4$Harbin Institute of Technology(Shenzhen), $^5$Beijing Institute of Technology \\ \texttt{hzmyouxiang@gmail.com}, \texttt{jzhx@pku.edu.cn}, \texttt{wk258730@alibaba-inc.com}, \\ \texttt{ljt@suda.edu.cn}, \texttt{victor\_fengss@foxmail.com}, \texttt{maoxl@bit.edu.cn}
  }

\begin{document}
	\maketitle
	\begin{abstract}
		Currently, human-bot symbiosis dialog systems, e.g., pre- and after-sales in E-commerce, are ubiquitous, and the dialog routing component is essential to improve the overall efficiency, reduce human resource cost, and enhance user experience. Although most existing methods can fulfil this requirement, they can only model single-source dialog data and cannot effectively capture the underlying knowledge of relations among data and subtasks. In this paper, we investigate this important problem by thoroughly mining both the data-to-task and task-to-task knowledge among various kinds of dialog data. To achieve the above targets, we propose a Gated Mechanism enhanced Multi-task Model (G3M), specifically including a novel dialog encoder and two tailored gated mechanism modules. The proposed method can play the role of hierarchical information filtering and is non-invasive to existing dialog systems. Based on two datasets collected from real world applications, extensive experimental results demonstrate the effectiveness of our method, which achieves the state-of-the-art performance by improving 8.7\%/11.8\% on RMSE metric and 2.2\%/4.4\% on F1 metric.
		
		%The tailored model structure for dialog routing can further capture the underlying mutual correlation between multiple subtasks
		
		%usually triggered by asking user to click a button or say something like 'transfer to human', with a less-friendly user experience. Thus, a more intelligent and reliable dialog routing component is essential to improve the overall system's efficiency and reduce human resource cost. However, it is still less-studied in the human-bot hybrid dialog systems. In this paper, we investigate a real-world problem of how to transfer the conversation among bot and human agents with deep learning methods. Technically, we first identify the problem as a multi-task decision making process which includes text regression and text classification. Then within a multi-task learning framework, a novel dialog encoder and a multi-gated mechanism are proposed to fuse the rich heterogeneous dialog information 
		%as well as to better capture the underlying mutual correlation between the two tasks, respectively. Experiments on two real-world datasets show that our method outperforms the state-of-the-art counterparts with 2.2\%-4.4\% and 8.7\%-11.8\% absolute improvement on two metrics.
	\end{abstract}
	
	\section{Introduction}
	
	Traditionally, a lot of human resource cost is spent on supporting the calling/online centers for customer care, such as pre-and after-sales for E-commerce and banking. With the rapid development of AI techniques for dialog systems, various bot agents have been deployed in those scenarios to reduce parts of human workload.  Both the bot and human agents constitute a new and practical symbiosis dialog system which can keep a balance between service quality and human resource cost~\cite{oraby2017may}.
	
	\begin{figure}
		\centering
		\includegraphics[width=0.5\textwidth]{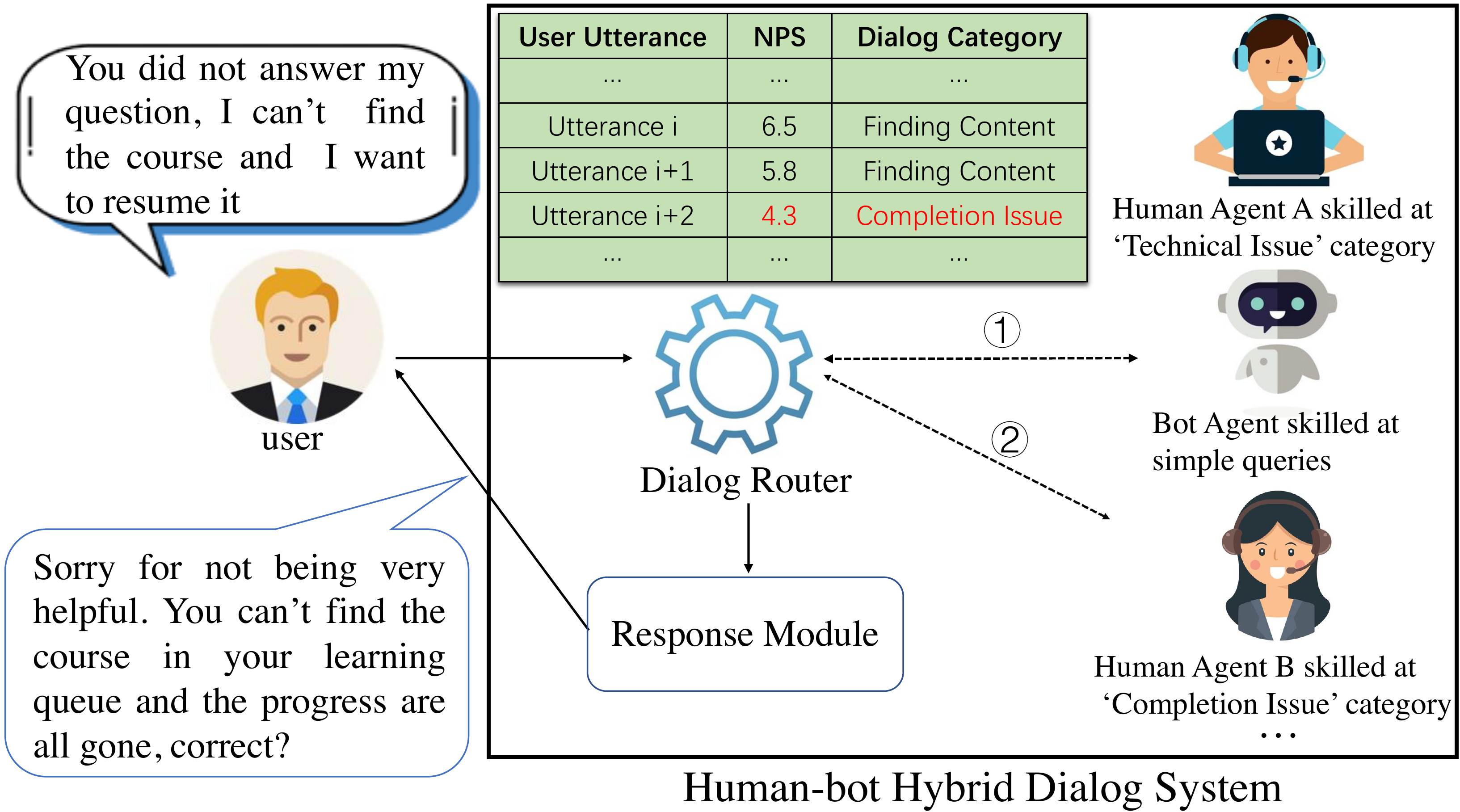}
		\caption{An example of dialog routing in human-bot symbiosis dialog system. \label{fig:example}}
		\vspace{-3ex}
	\end{figure}
	%where previously are occupied by expensive human agents.
	%fuzhu to keep business value
	
	%fully leverage learning ability to improve the performance
	
	In such a human-bot symbiosis environment, it is commonly seen that users' needs continuously change, or sometimes they are dissatisfactory with the current servicing agents.
	%Such mismatching often happens. 
	They would request the dialog system to replace the agents with more qualified ones, because usually each bot or human agent is trained for specific domains. For example, some agents are good at handling after-sales complaints, while some others would do well in providing technical guidance. To make the whole system operate efficiently, a dialog routing component is necessary to take charge of when to transfer the conversation and which domain of agents is suitable to provide service. 
	
	To illustrate the process of dialog routing task more clearly, Figure~\ref{fig:example} shows an example. At the beginning, the user’s query is operated by the bot agent through Flow \textcircled{\small{1}}, and the dialog router would continuously monitor the user's satisfactory degree by inferring Net Promoter Score (NPS) and current dialog category. When the inferred NPS (4.8) is lower than a hand-crafted threshold (5) at the $i$th dialog turn, the router chooses to switch the dialog to one of human agents skilled at the inferred category via Flow \textcircled{\small{2}}. Here human agent B is good at the category of `Completion Issue' dialogs. From this picture, we could realize that how to implement the accurate matching between user's needs and skilled agents is greatly correlated with the whole system's efficiency and user experience.
	
	Currently, most existing deployed systems are heuristic and support to trigger the dialog routing when users actively click a button or say/type something like `transfer to human'. This method is less user-friendly and cannot fully leverage the ability of  contemporary deep leaning techniques. Recently, some learning based methods are proposed. Yu et al. well defines the dialog routing as text regression and classification problems, and proposes a learning network with CNN and RNN modules to encode dialog data~\cite{Yu2020}. However, it only models the token-level utterance data and misses the information from other kinds of dialog data\footnote{Other kinds of dialog data include speaker roles, utterance-level sequence, intent information, etc.}; Meanwhile the model structure is shared by the two tasks (in single-task way) which cannot capture the underlying knowledge within data-to-task (intrinsic) and task-to-task (extrinsic) relations. In this paper, we investigate the dialog routing task from the multi-task learning perspective to further capture the intrinsic and extrinsic information.

	Intuitively, the decision making process of dialog router could be decomposed to two subtasks, NPS prediction and dialog category classification. We adopt the same NPS prediction task as \cite{Yu2020} since the supervised data can be obtained without any human-labeling effort. Furthermore, we observe that the data-to-task (intrinsic) and task-to-task (extrinsic) information should be leveraged to make better routing performance.  On the one hand, we can argue that the more accurate NPS is predicted, the more precise the dialog category is classified, and vice verse. On the other hand, in addition to the token-level utterance data, other kinds of dialog data can also affect the final task performance. For example, if a user utterance is about the intent of `ask\_technical\_problem', the intent information is an obvious indicator that the user should be served by the agent who is skilled at `Technical Issue' category. Note that the categories of agents and dialog intents of users are often different, and usually the former is much less than the later.
	
	%Technically as to the multi-task learning problem, there are two-fold challenges: 1) dialog data modeling. As known, dialog data contains a lot of kinds of data formats, e.g., token sequence of an utterance, utterance sequence of a dialog session, utterance role, intent of each utterance (known information). It is difficult to fuse the rich textual information. 2) correlation learning between task to task and task to data. Intuitively we can assume that the more accurate NPS is predicted, the more precise the dialog category is classified, and vice verse. Moreover, intent data also affects dialog category's prediction, e.g., if an utterance of a dialog contains `I\_have\_a\_technical\_problem' intent, this dialog probably belongs to `Technical Issue' dialog category. Therefore, how to fully learn the underlying correlation between the NPS and dialog category tasks and relationship between intent and dialog category task are crucial to improve the performance.
	
	To achieve the above motivation, in this paper, we propose a novel Gated Mechanism enhanced Multi-task Model (G3M). Firstly, the model extends the BERT encoder to encode various kinds of dialog data in a hierarchical way. Moreover, two modules of gated mechanism are proposed to explicitly model the data-to-task and task-to-task information under multi-task learning framework. Another advantage of G3M is its good compatibility, such that it can be easily integrated into existing human-bot dialog systems in a plug-in manner. We conduct various experiments on two datasets collected from the real world. The results demonstrate our method can achieve the state-of-the-art performance on both tasks, and the ablation experiment proves our model is effective to simultaneously capture the intrinsic and extrinsic information.

	In summary, this paper's contributions include:
	\begin{itemize}
		%\setlength{\itemsep}{0pt}
		%  \setlength{\parskip}{0pt}
		%  \setlength{\itemindent}{0em}
		
		%   \item The application-oriented problem of dialog routing from the perspective of multi-task learning is identified. How to model dialog data and capture the correlation between task to task and data to task are the two key problems.
		%   \item a Multi-task Multi-gated Mechanism (M3) on BERT to make NPS prediction and topic classification.
		%   We propose a Multi-task Multi-gated Mechanism (M3) on BERT for dialog information fusion and multi-task correlation learning.
		%   \item Two real-world datasets are collected and experiments show that our model can greatly outperform the state-of-the-art baselines.
		%\item We identify a novel application-oriented problem of dialog routing, which is a multi-task learning problem. It targets to model dialog data and capture the correlation between task to task and data to task.  
		\item We argue that both the data-to-task and task-to-task information are important to achieve better dialog routing. Hence, we propose a new multi-task learning solution, called Gated Mechanism enhanced Multi-task Model (G3M), to implement the motivation.
		%\item We propose a Gated Mechanism enhanced Multi-task Model on BERT to make NPS prediction and dialog classification.
		%The proposed model is able to address dialog information fusion and multi-task correlation learning.
		\item We extend the BERT encoder to encode various kinds of dialog data in a hierarchical manner, and develop two modules of gated mechanism to explicitly model the data-to-task and task-to-task information.
		\item We conduct extensive experiments on two real world datasets. The results prove our model's effectiveness, which can achieve the state-of-the-art performance.
	\end{itemize}
	
	% A novel pre-trained joint learning model with  knowledge-gated mechanism has been proposed to solve tasks simultaneously. The model not only can explicitly capture the relationship between tasks and added it to model learning via knowledge-gated mechanism and implicitly utilize the semantic correlation between tasks via pre-trained joint learning manner, but also learn an end-to-end solution from heterogeneous dialog information.
	%The model not only can explicitly and implicitly consider the relationship between tasks via knowledge gating results and joint learning manner respectively, but also learn an end-to-end solution from heterogeneous dialog information.
	\begin{figure*}[t]
		\centering
		\includegraphics[width=0.88\textwidth]{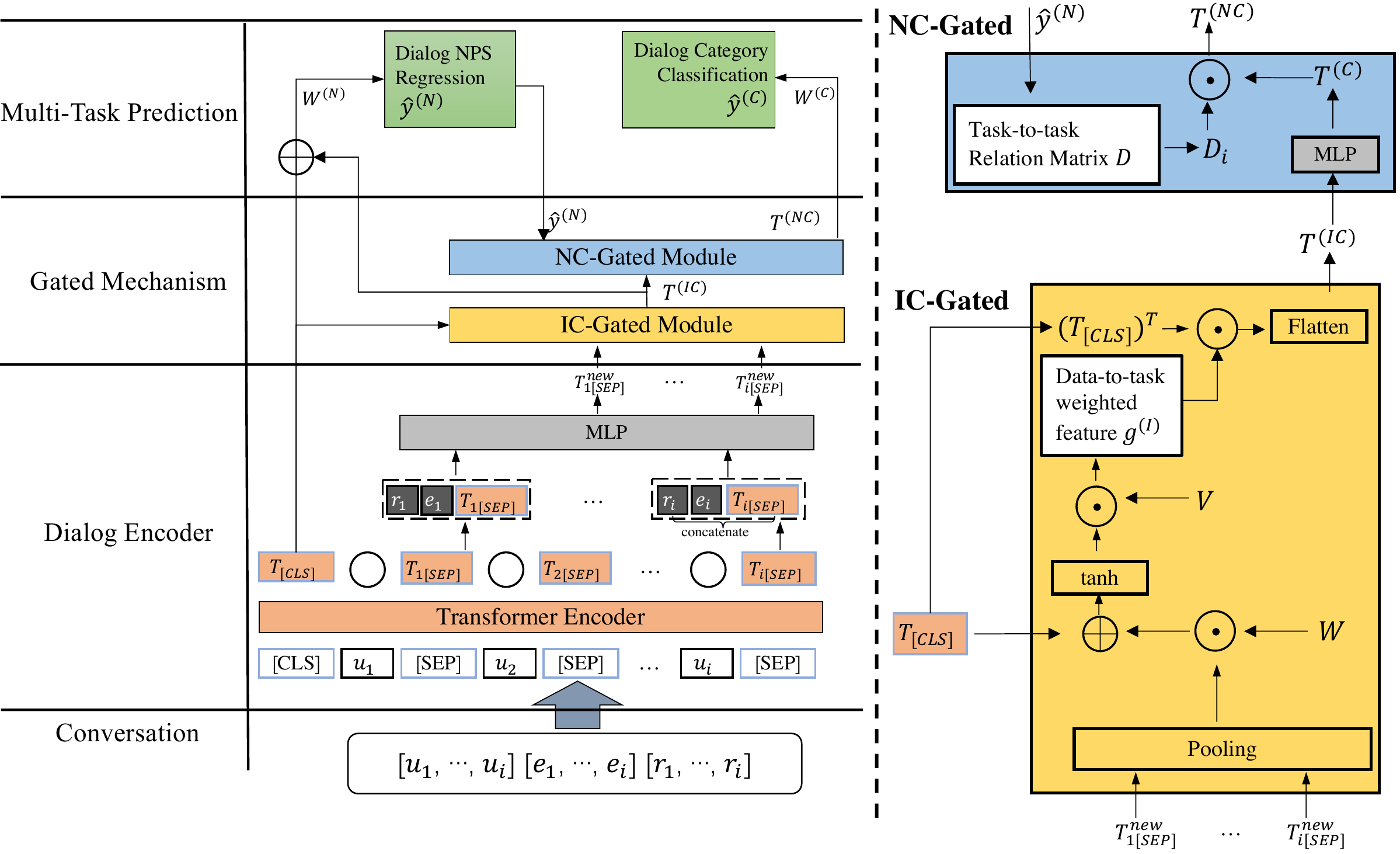}
		\vspace{-1ex}
		\caption{
			%   Model's architecture with zoomed in IC-Gated module (yellow part) and NC-Gated module (blue part).
			Architecture overview of Gated Mechanism enhanced Multi-task Model with zoomed in NC-Gated (right top blue part) and IC-Gated (right down yellow part) modules.
			%   Model architecture as left part shows. The detail design for NC-Gated and iC-Gated are shown in right top (blue) and down part (yellow) respectively. The detail of statistical distribution and important intents list are described in Section \ref{sec:model}.
		}
		\label{fig:model}
		\vspace{-2ex}
	\end{figure*}
	\section{Related Work} \label{sec:related-work}

	Compared with other dialog system tasks, there are not many related work on dialog routing study. 
	Recently, a learning network with attention based CNN and RNN is proposed~\cite{Yu2020}. It mainly models the token-level dialog data and is a single-task model, i.e., the NPS regression and dialog classification tasks share the same model structure except the last inference layer. This method cannot sufficiently leverage the underlying knowledge within subtasks and various kinds of dialog data. Another similar work is a demo system for dialog transition~\cite{huang2021dialog} which tries a vanilla multi-task learning method (i.e., a dialog encoder tailed with two prediction subtasks). However, it still hardly fully utilizes the knowledge among dialog data and subtasks. Different from them, we further investigate the problem by simultaneously modeling data-to-task and task-to-task information, achieving a much advanced performance.%Our method is inspired by them but further investigates the problem by simultaneously modeling data-to-task abd task-to-task information, achieving a much advanced performance.
	
	The dialog routing task is fundamentally decomposed to two subtasks, dialog NPS regression and dialog classification, which feature encoding various kinds of dialog data and have been separately studied by researchers from the NLP community. The representative research studies include CNN-based models \cite{Kim2014}, RNN-based models \cite{Wang2018}, regression-based models~\cite{Dereli2019,ngo2014influence} and deep bi-directional transformers model with pre-training \cite{Devlin2018,cohan2019pretrained}. From these works, we learn that dialog data encoder is the common key module to make sure good performance on final tasks~\cite{Dereli2019}. Those methods have not been demonstrated to be effective in the multi-task situation of dialog routing, without thoroughly mining the relations among data and subtasks.

	%One of the keys of text regression and classification in our dialog routing task is how to well encode the heterogeneous dialog information~\cite{Dereli2019}. 
	%Encoding methods for general text classification have been largely studied, such as CNN-based models \cite{Kim2014}, RNN-based models \cite{Wang2018} or a pre-trained of deep bi-directional transformers model \cite{Devlin2018,cohan2019pretrained}, dialog category classification is actually a little different with those. The dialog data in our situation has a various kinds of formats, including token sequences, sentence sequences, roles, intents and etc. They should be well fused by elaborating a situation-specific information fusion model.
	%There is a very similar work to ours~\cite{Yu2020} and we have the same target for dialog routing. However it is a single task and we argue that multi-task modeling like ours can achieve much better performance in real situations.
	
	Recently, multi-task learning on dialog data has been studied and proven successful, such as joint slot filling and intent prediction~\cite{goo2018slot}, dialog act sequence labeling~\cite{Kumar2018}, and dialog response generation~\cite{naacl21}. These studies suggest that dialog data contains rich information and multi-task learning could dig out the underlying knowledge among data and subtasks. Moreover, among multi-task learning methods, multi-gated mechanism is widely utilized for information remembering and filtering~\cite{Yubo2018,Xiao18,du2019knowledge}, which is much suitable for tidying dialog data. Inspired by the above ideas, we propose a novel model to combine them together in our solution for further improving the performance of dialog routing task.%We are inspired by all the above ideas and propose a novel model to combine them together in our solution for improving the performance of dialog routing task.

	\section{Problem Definition}
	
	We define the investigated problem formally in this section. In our task, we use the available data in a conversation session as inputs, including an utterance list $U = \{u_1,...,u_L\}$, a speaker role list $R = \{r_1,...,r_L\}$, and an intent list $I = \{e_1,...,e_L\}$, where $u_i$ is an utterance, $r_i$ indicates the role of the $i$th utterance (user or agent), $e_i$ is known intent of the $i$th utterance, and $L$ is the number of utterance sequences in a conversation session. Each utterance has a token-level sequence $u_i = \{w_{i1},..., w_{iM}\}$, where $w_{ij}$ means the $j$th token in the $i$th utterance and $M$ is the length of the token sequence.
	%with zero paddings or truncation if it is less then or larger than $M$. 
	Here we can leverage the intent information since we use an intent extraction tool\footnote{More information can be referred in Section Experiments.}. Our model is also scalable to capture more dialog data, such as entity and dialog act information if they are available. We adopt the intent information in our solution based on the intuitive consideration of data-to-task motivation.
	
	%We define our problem formally in this section. In our situation, the available data in a conversation session includes a utterance list $U = \{u_1,...,u_L\}$, a role list $R = \{r_1,...,r_L\}$ and an intent list $I = \{I_1,...,I_L\}$ corresponding to each utterance $u_i\in U$, where $L$ is the number of utterances in a conversation session and $r_i$ indicates the side of the $i$-th utterance. Each utterance has a token sequence $u_i = \{w_i^{(1)},..., w_i^{(M)}\}$, where $w_i^{(j)}$ is the the $j$-th token in the $i$-th utterance and M is the length of the token sequence which is a fix preset value aligned with pre-trained encoder.
	%with zero paddings or truncation if it is less then or larger than $M$. 
	%Note that although other more kinds of dialog-related information, e.g. entity
	%information and dialog act and etc., can be modeled together, this paper only considers those which are relevant to the dialog routing task.
	
	%The intent information is prior knowledge while we leverage a tool\footnote{anonymous under review} to do tagging. Each $I_i$ is a one-hot like vector and the vector dimension is the same with the number of preset different intent categories. In our experiments, we define 66 and 28 intent classes for two scenarios and will introduce the details in the following Experiments Section. 
	For the outputs of our task, at each dialog turn $i$, our model can make predictions of NPS $y_i^{(N)}$ and dialog category $y_i^{(C)}$. In summary, given $U_i = \{u_1,...,u_i\}$, $R_i = \{r_1,...,r_i\}$ and $I_i = \{e_1,...,e_i\}$, we formally state the problem as follows:
	
	\begin{equation}
		y_i^{(N)}, y_i^{(C)}=f(U_i, R_i, I_i).
		%Input: [U_i, R_i, I_i] \longrightarrow Output: [y_i^{(N)}, y_i^{(C)}]
	\end{equation}

	\section{Proposed Method}\label{sec:model}
	The proposed model generally contains three parts: (1) Dialog Encoder, (2) Gated Mechanism, and (3) Multi-Task Prediction. Figure~\ref{fig:model} shows the overview of model's architecture, which will be introduced next. % and we introduce them in the following.
	
	\subsection{Dialog Encoder}
	
	% To model the complex format of dialog data, it should be elaborate to organize the data and design an efficient method. 
	
	% Inspired by the success of pre-trained models, e.g. BERT \cite{Devlin2018} and 
	To encode the various kinds of dialog data, we leverage and extend the BERT encoder. As presented in Figure~\ref{fig:model}, given an utterance sequence $\{u_1,...,u_i\}$, firstly all the utterances are flattened and concatenated to a long token sequence. Then we add a [CLS] token at the beginning and insert [SEP] tokens between any two utterances. Similar to BERT encoder that embeds the whole token sequence by Transformer structure, we get the representation of the $i$th  [SEP] token, $T_{i[SEP]}$, and regard it as the representation of the $i$th utterance. The representation of head token, $T_{[CLS]}$, would be used in the following Gated Mechanism part as a kind of context information which encodes all the utterances.
	
	With the utterance representations of $T_{i[SEP]}$, the role $\{r_1,...,r_i\}$ and intent $\{e_i,...,e_i\}$ information (one-hot vectors) are concatenated to their corresponding utterance representations.  After a MLP operation, we can obtain the final representation of each utterance by $T^{new}_{i[SEP]} = MLP(r_i \oplus e_i \oplus T_{i[SEP]})$.
	
	%To encode our rich dialog information, we propose an information fusion model. As present in \autoref{fig:model}, given $\{u_1,...,u_i\}$, we separate each utterance by inserting a delimiter [SEP], and [CLS] is added at the beginning. The operation is similar to BERT but with a longer sentence list. After operating by Transformer component of BERT, the vector $T_{i[SEP]}$ is leveraged to represent the $i$-th utterance and $T_{[CLS]}$ is used to represent all the utterances as context information. At last in the Dialog Encoder, the concatenated representation $T^{(rI)}_{i[SEP]}$, which includes the role information $\{r_1,...,r_i\}$, intent information $\{I_i,...,I_i\}$ and each utterance's representation $T_{i[SEP]}$ to generate the final enhanced representation $T^{new}_{i[SEP]}$ for each utterance.
	%is operated by a multi-layer perceptron (MLP) and $T^{new}_{i[SEP]} = MLP(T^{r}_{i[SEP]}\oplus I_i)$ is the final representation for each utterance.
	
	%\begin{equation}
	%	\begin{aligned}
	%		T^{(rI)}_{i[SEP]} = r_i \oplus I_i \oplus T_{i[SEP]} \\
	%		T^{new}_{i[SEP]} = MLP(T^{(rI)}_{i[SEP]})
	%	\end{aligned}
	%\end{equation}
	
	Our method to encode dialog data is inspired by the Sequential Sentence Classification (SCC) model~\cite{cohan2019pretrained} which also is based on BERT encoder and organizes the dialog data in the hierarchical manner. However, our model is different from SCC model in two aspects. Firstly, the representation of head token [CLS] is additionally utilized as context information in the following modules. Secondly, our encoder can integrate the extra role and intent information.

	\subsection{Gated Mechanism}
	
	Besides the basic mechanism via sharing parameters of the encoder in vanilla multi-task learning paradigm, we leverage gated mechanism and propose two modules for dialog routing to better model the task-to-task and data-to-task information.
	%Beyond basic parameter sharing mechanism in general multi-task learning paradigm, as we introduced in the introduction, there are explicit mutual correlation between task to task and data to task in dialog routing scenario, we propose another adaptive novelty in our situation, the multi-gated mechanism to better model and memorize the mutual correlation between data to task and task to task explicitly.
	
	\subsubsection{IC-Gated Module}
	
	The \textbf{I}ntent-\textbf{C}ategory-Gated (IC-Gated) module is designed to model the relation between intent, role and dialog category (data-to-task) information. The right down yellow part in Figure~\ref{fig:model} illustrates the detailed module structure. 
	
	With the context representation $T_{[CLS]}$ and utterance representations $T^{new}_{i[SEP]}$, we extend a commonly used way~\cite{goo2018slot} to capture the relations between intent, role and dialog context by learning a weighted feature for data-to-task (i.e., intent, role to dialog category) modeling. Thus we obtain the weighted feature from various levels of the input dialog data after max pooling operation on the utterance sequence by the following equation:
	\begin{equation}\small
		\begin{aligned}
		%T^I = Pooling(T_{1[SEP]}^{new}, ..., T_{i[SEP]}^{new}) \\
		g^{(I)} &= V\cdot\tanh(T_{[CLS]}\\&+W\cdot Pooling(T_{1[SEP]}^{new}, ..., T_{i[SEP]}^{new})),
		\end{aligned}
		\label{eq:gI}
	\end{equation}
	where $V$ and $W$ denote the parameters to learn. 

	With the weighted feature $g^{(I)}$ that attends intent and role information on dialog context, the context information $T_{[CLS]}$ is used again to calculate the final weighted representation of dialog data with a flatten operation as the following equation\footnote{$Flatten(\cdot)$ function reshapes the input matrix into a one-dimension vector.}:
	%At last, we calculate the final output $T^{IC}$ of the IC-Gated module by a flatten operation on the global context and weighted feature as the following:
	\begin{equation}\small
		T^{(IC)} = Flatten(g^{(I)}\cdot (T_{[CLS]})^\mathrm{T}).
	\end{equation}
	
	The IC-Gated module can learn the representation from dialog data and meanwhile consider the related information between intent, role and dialog context, which would benefit the later dialog category classification task.
	%After the IC-Gated module we generate the $T^{IC}$ representation which is not only learned from context but also contained the mutual correlation between intent and dialog category explicitly. We will leverage $T^{IC}$ vector to predict the dialog routing tasks.
	
	% This module is inspired by a similar work~\cite{goo2018slot}, but we extend their work to our dialog situation and add a flatten operation to capture more context information in order to further build the relationship with local and global textual contents.
	
	\iffalse
	\begin{figure}
		\centering
		\includegraphics[width=0.3\textwidth]{figure/ic-gate}
		\caption{Illustration of iC-gated mechanism. \label{fig:ic-gate}}
	\end{figure}
	\fi

	\subsubsection{NC-Gated Module}
	%This section describes the proposed NC-gated mechanism illustrated in the blue part of \autoref{fig:model}.
	The \textbf{N}PS-\textbf{C}ategory-Gated (NC-Gated) module is designed to further model the underlying relation between subtasks (task-to-task). The right top blue part in Figure~\ref{fig:model} shows the structure detail.
	
	%Beyond sharing parameters by multi-task framework to learn inexplicit relationship between NPS prediction and dialog classification task, we also proposed the \textbf{N}PS-\textbf{C}ategory-Gated (NC-Gated) module to learn explicit correlation between two tasks. The detailed design is shown by the blue part in \autoref{fig:model}.
	
	We set a task-to-task relation matrix $D\in\mathbb{R}^{d^N \times d^C}$ to preserve the empirical distribution of co-occurrence between NPS and dialog categories among dataset with min-max normalization \cite{Singh2015}, where $d^N$ and $d^C$ are the numbers of NPS intervals and dialog categories respectively. We partition the NPS ranging from 0 to 10 into ten intervals. Then with the representation from IC-Gated module $T^{(IC)}$, the output representation of NC-Gated module is calculated by combining a MLP layer and the task-to-task relation matrix $D$:
	%In the NC-Gated module, we set a matrix $D \in \mathbb{R}^{d^N \times d^C}$ for preserving the distribution of co-occurrence between different NPS and category pairs in the training dataset with normalized them according to min-max normalization \cite{Singh2015}, where $d^N$ and $d^C$ are the number of different NPS intervals and dialog category types respectively. $D_i$ is denoted as the $i$th row which means a one-hot-like representation corresponding to the $i$th NPS. The $i$th NPS means we averagely partition the range of NPS from 0 to 10 with ten intervals. Then $T^{(IC)}$ is operated by a MLP layer and its output $T^C$ is used with $D_i$ to calculate the final output of NC-Gated module. $T^{(NC)} \in \mathbb{R}^{d^C}$ is calculated by \autoref{eq:ic} as input fo the dialog category classification prediction.
	
	\begin{equation}\small
		%	\begin{aligned}
		%		T^C = MLP(T^{(IC)}) \\
		T^{(NC)}= D_{i}\odot MLP(T^{(IC)}),
		\label{eq:ic}
		%	\end{aligned}
	\end{equation}
	where $\odot$ is Hadamard product to measure the similarity. $D_i$ is the $i$th row vector which means a representation of the $i$th NPS interval corresponding to every dialog category, and it depends on the predicted NPS. The reason that uses NPS to query the matrix instead of using predicted dialog category to query NPS, is that we empirically find the effect from the NPS regression task is more significant than the dialog category classification task. %The reason of why we use NPS to query the matrix, rather than using predicted dialog category to query NPS, is that by our experiment we find the effect from the NPS regression task is much significant than the dialog category classification task, although the effect from the contrary direction is still present. 
	
	Note that except for the multi-task learning paradigm that can capture the relation between tasks via parameter sharing, we use the relation matrix $D$ as prior knowledge (or attention weights) and design the NC-Gated module for better representation learning of dialog data by further capturing the task-to-task information.

	\subsection{Multi-Task Prediction}
	
	Dialog routing has two subtasks, namely NPS regression and dialog category classification. For the former task, the enhanced representation $T^{(IC)}$ learned by the IC-Gated module from data-to-task information and the dialog context representation $T_{[CLS]}$ are concatenated together to make the prediction of NPS. The predicted NPS $\hat{y}^{(N)}$ can be defined as:
	%In our situation, there are two tasks, NPS regression and dialog category classification. 
	%The first task is NPS regression prediction. The model calculates the NPS belongs to the current dialogue. The relation between NPS and intent is modelled and influenced by $T^{(IC)}$. The predicted NPS $\hat{y}^{N}$ can be denoted as:
	\begin{equation}\small
		\hat{y}^{(N)} = W^{(N)}(T_{[CLS]}\oplus T^{(IC)}) + b^{(N)}.
	\end{equation}
	$W^{(N)}$ and $b^{(N)}$ are parameters to learn.
	%where $g(\cdot)$ is a regression function to map from z-score in \autoref{eq:z-score} to the actual value of predicted NPS.
	
	The dialog classification task uses the enhanced representation $T^{(NC)}$ as input, which is learned by the NC-Gated module from task-to-task information. The predicted dialog category $\hat{y}^{(C)}$ is calculated by a softmax function:
	%The second task is dialogue category classification task. The classifier predicts the category means which category the dialogue
	%talks about belongs to. It's a multiple classes classification problem. We use a softmax function to calculate each category’s probability and choose the maximum as the
	%predicted class. Assume that we have $d^C$ different category classes and the predicted category $\hat{y}^{C}$ can be denoted:
	\begin{equation}\small
		\hat{y}^{(C)} = Softmax(W^{(C)}T^{(NC)} + b^{(C)}).
	\end{equation}
	%where $f(\cdot)$ is a softmaxt function.
	$W^{(C)}$ and $b^{(C)}$ are parameters to learn.
	
	\subsection{Model Training}
	To jointly train our model from the two tasks together under the multi-task learning paradigm, we design a loss function by combining a mean squared loss for NPS regression and a cross-entropy loss for dialog classification. To avoid over-fitting, $L_2$ regularization terms are adopted. The overall objective function is defined as:
	
	%By combining these two tasks, we design a loss function for jointly training the model by adding the two tasks' loss functions. For the NPS regression task, we set a mean squared loss, while for the classification task, we set a cross-entropy loss function. The overall objective is to minimize the following loss function:
	\begin{equation}\small
		%\begin{split}
		\begin{aligned}
			\mathcal{L} = &\underbrace{\frac{\alpha}{N}\sum_{k=1}^N[\hat{y}_k^{(N)} -y_k^{(N)}]^2}_{\text{NPS regression task}}-\underbrace{\frac{\beta}{N}\sum_{k=1}^{N}\sum_{y}y_k^{(C)}\log(\hat{y}_k^{(C)})}_{\text{dialog classification task}}\\&+\underbrace{\frac{\gamma}{2}(\rVert W^{(N)} \rVert_2^2+\rVert W^{(C)} \rVert_2^2)}_{\text{regularization terms}},
		\end{aligned}
		%\end{split}
	\end{equation}
	
	where $N$ is the number of all training samples, $\hat{y}_k^{(N)}$ and $\hat{y}_k^{(C)}$ are the inferred NPS and category labels, and	$y_k^{(N)}$ and $y_k^{(C)}$ are the ground truth.
	%The first term is log-likelihood cross-entropy loss function for category prediction task, the second one is mean square error loss function for NPS prediction task, and the last one is a 
	%We add a regularization term to avoid over-fitting. 
	$\alpha$, $\beta$ and $\gamma$ are hyper-parameters to balance the weights of terms.
	
	\iffalse
	y_hat^C_k is predicted category vector for k-th sample and y^C_k \in R d_C (0,1) is a vector with only one non-zero element that represents its true category label.
	y_hat^N_k is predicted NPS value for k-th sample and y^N_k is ture NPS value
	Similarly, y^I-{kj} is predicted intent vector from the model of jth utterance for k-th sample and y_hat^I_{kj} \in  is a vector with only one non-zero element that represents its true intent label.

	%$y_{kt}^{C}$ is the ground truth (0 or 1) for $d_{t}^{C}$, where $d_t^C$ is $t$-th correct label of \emph{Category} in $k$-th sample,
	the true \textit{NPS} is $y_k^{N} in $ for $k$-th sample, $\hat{y}_k^{N}$ is $k$-th predicted \emph{NPS}, $\hat{y}_{kj}^I$ is predicted intent of $j$-th utterance in $k$-th sample.
	% and $y_t^{I}$ is the ground truth (0 or 1) for $d^{I}_{t}$, where $d_t^I$ is $t$-th correct label of intent.
	The first term is for category prediction, the second is for NPS prediction and the last one is for sequential intent classification respectively. The $\alpha$, $\beta$ and $\gamma$ are hyper-parameters to control the contributions of three parts.
	\fi

	\section{Experiments}

	%In this section, we report the effectiveness of our model on two real-world datasets for the two tasks: NPS prediction and category classification.

	\subsection{Corpus}
	
	To our knowledge, there is no public corpus contains both the NPS and category labels, we collect two corpora from the real-world dialog systems to evaluate our method.
	%Consider our application-oriented problem, we collect two private real-world dialog datasets for evaluation. 
	
	\textbf{\emph{Learning Corpus}} is collected from an online education platform,
	%\footnote{URL will be public when this paper is accepted.}
	where the dialog system for customer care is constituted of 7 dialog categories with all human agents: Badge Issue, Completion Issue, Finding Content, System Help, Technical Issue, Ticket Status, and Other.
	%The dialogs logged by the system are mainly about course learning, course information, network issue and so on. 
	%In total, we have 22897 sessions of conversation and 7 human agent categories in this dataset, including Badge Issue, Completion Issue, Finding Content, System Help, Technical Issue, Ticket Status, Other. We also define 66 types of intent in this dataset.
	
	\textbf{\emph{MacHelp Corpus}} is collected from an after-sale technical support platform
	%\footnote{URL will be public when this paper is accepted.}
	for users who have troubles with their Mac laptops, where the dialog system is constituted of 4 dialog categories for human agents: Login Problem, Apps Issue, Raise Tickets, Account Issue.
	%We collect 37944 sessions of conversations and 4 category: Login Problem, Apps Issue, Raise Tickets, Account Issue.
	%The MacHelp dataset contains 28 types of intent.
	
	%One is from an online employee training platform\footnote{URL will be public after published} and is called Learning dataset. The other is from an after-sales platform\footnote{URL will be public after published} which provides technical support for Mac laptop users and is called MacHelp dataset. 
	
	% The statistics of the two datasets and the distribution and explanation of the Category are shown in \autorefs{table:statistics} and \ref{table:distribution_yl} respectively.
	
	% The Learning dataset has 7 topic categories: Badge Issue, Completion Course, Finding Content, System Help, Technical Issue, Ticket Status, Other. The MacHelp dataset has 4 topic categories: Login Problem, Apps Issue, Raise Tickets, Account Issue. 
	\subsection{Corpus Preprocessing}
	
	For the two raw corpora, we preprocess them into a consistent format for model training and testing. Table~\ref{table:example} shows an example of the data structure of dialog session. In a dialog session, we have an utterance sequence, a role sequence to indicate the speaker role of each utterance (0 means user and 1 means agent), and an intent sequence labeled by using an online tool which is separately trained from supervised information for the domain of customer care\footnote{https://www.ibm.com/products/watson-assistant. The intent detection accuracy is about 92\% in additional assessment.}. We use the off-the-shelf extraction tool and obtain 66 and 28 types of intents on the two corpora respectively.
	%We preprocess the two datasets into consistent format. For each historical dialog session, we tag the session with a human-labelled category, and a NPS label which is from the user's feedback. Moreover, each utterance corresponding to a dialog session has a role label, e.g. 0 means user and 1 means agent. The each utterance's intent label is by using an existing tagger\footnote{URL will be public after published, the accuracy of intent detection is about 92\%.}.
	%\footnote{https://www.ibm.com/cloud/watson-assistant/} 
	%and then with manual double checking.
	% The intents are labelled by leveraging a commercial conversational system \footnote{https://www.ibm.com/cloud/watson-assistant/} with human double confirm. 
	%\autorefs{table:example} shows an example of the data structure and \autorefs{table:statistics} shows some statistic information of the two datasets.
	
	Within a dialog session, we partition the data to several training and testing samples. At the $i$th dialog turn, a sample is defined as $S_i=\{\{u_1,...,u_i\}, \{r_1,...,r_i\}, \{e_1,...,e_i\}, y_i^{(N)}, y_i^{(C)}\}$. Ground truths are obtained by the following rules:
	\begin{itemize}
		\item The NPS of $u_{i+1}$ is regarded as the label (i.e., $y_i^{(N)}$) for NPS regression task.
		\item The agent category of $u_{i+1}$ is regarded as the label (i.e., $y_i^{(C)}$) for dialog classification task.
		\item If there is no feedback provided by user on the session, our task degrades to gated-mechanism single-task dialog classification.
	\end{itemize}
	
	Note that although the ground truths of NPS and category label are naturally obtained from the users' feedback and human agent category, to guarantee the data quality, three humans are invited to review the correctness of all the labels, and we only adopt the samples agreed by all of them. As a result, we obtain about 90\% valid sessions from the corpora. Table~\ref{table:statistics} shows the statistic information of the final two processed datasets.
	
	%More detailed processing operations include each dataset is randomly split into training/validating/testing sets in the proportion of 8:1:1. Since the BERT~\cite{Devlin2018} encoder only supports a length of token sequence up to 512, to handle the longer dialog data, we
	%remove greeting or closing utterances at the beginning and
	%the end of a conversation. Finally, we get about 90\% valid dialogs whose length is less than or equal to 512\footnote{We can pre-train a new BERT encoder with a longer input length (greater than 512) to handle the longer sequence scenario, we leave it as a future work.}.
	
	% The range of NPS is from 0 to 10. 
	The real NPS is from user's feedback by clicking or typing a score, and the two corpora have different NPS ranges (one is an integer in [0,10] and the other is a decimal in [0,5]). Therefore we use zero-mean normalization (z-score)~\cite{Singh2015} method to normalize the scores with a mean of 0 and a standard deviation of 1 as the following: $\text{z-score} = \frac{x_i-\overline{x}}{\sigma}$, 
	%We have almost 47\% and 56\% of total data are labelled with NPS value for Learning and MacHelp dataset respectively. To get better NPS regression results, zero-mean normalization (z-score)~\cite{Singh2015} method is leveraged to normalize
	%the NPS with a mean of 0 and a standard deviation of 1.
	% The NPS that originally ranges from 0 to 10 is normalized to standard Gaussian distribution by zero-mean normalization (z-score) \cite{Singh2015} method.
	%\begin{equation}\small
	%	\text{z-score} = \frac{x_i-\overline{x}}{\sigma},
	%	\label{eq:z-score}
	%\end{equation}
	where $x_i$ is NPS, $\overline{x}$ is the mean value and $\sigma$ is the standard deviation. By our statistics, we have about 47\% and 56\% of dialog sessions that have valid NPS labels for Learning and MacHelp datasets respectively, which makes the dialog routing become a much more challenging task in the real world.

	\begin{table}[]
		\small
		\begin{center}
			\scalebox{0.8}{
				\begin{tabular}{|l|l|l|}
					\hline
					\textbf{role}          & \textbf{utterance}                                                                                                                                    & \textbf{intent}   \\ \hline
					1 (agent) & \tabincell{l}{Hello, how may I assist you? }                                                                                                                         & greeting          \\ \hline
					0 (user) & \tabincell{l}{Hello   }                                                                                                                                              & greeting          \\ \hline
					0 (user) & \tabincell{l}{I am looking some training which will \\help me to learn more about Github.        }                                                                     & request \\ \hline
					1 (agent) & \tabincell{l}{In Your Learning webpage, you have \\a search field, you can type what you \\want to learn about, and it will bring \\you all courses related to that topic.} & guidance \\
					%\multicolumn{1}{|c|}{} & \multicolumn{1}{c|}{...} & \\ 
					\hline
					\multicolumn{3}{|l|}{\textbf{User NPS}: 9.0\quad \textbf{Dialog Category}: Finding Content} \\ \hline
			\end{tabular}} \caption{Data format example of a dialog session.} \label{table:example}
		\end{center}
		% \vspace{-2ex}
	\end{table}
	
	\begin{table}
		\small
		\centering\vspace{1mm}\scalebox{1}{
			\begin{tabular}{|c|c|c|}
				\hline
				&{Learning}&{MacHelp}\\
				\hline
				avg. tokens per utterance & 10.19 & 10.29 \\
				avg. turns per dialog session & 22.23 & 30.43 \\
				\# of total tokens & 66,031 & 154,023 \\
				\# of total dialog sessions & 22,897 & 37,944 \\
				\# of dialog categories & 7 & 4 \\
				\# of intents & 66 & 28 \\
				\hline
		\end{tabular}}  \caption{Statistics for Learning and MacHelp datasets} \label{table:statistics}
		\vspace{-3ex}
	\end{table}
	
	%\subsection{Measurement}
	%For the NPS regression task, root mean squad error (RMSE) is employed to measure the error, and the smaller RMSE means the better performance.
	%\begin{equation}
	%	\text{RMSE} = \sqrt{\frac{\sum_i(T_i-P_i)^2}{N}}
	%\end{equation}
	%where N is number of samples, $T_i$ and $P_i$ are true value and predict value respectively.
	
	%For the dialog category prediction task, which can be treated as a multi-class classification problem and the data is unbalanced, so the metric we employ micro F1 score (micro-F1), and the larger micro-F1 the better performance.
	
	%\begin{equation}
	%	\text{micro-F1} = \frac{\sum_c{TP_c}}{\sum_c{TP_c}+\sum_c{FP_c}}
	%\end{equation}
	%where c is the class label, TP is true positive and FP means false positive. %In micro-averaged method, recall equals precision.

	\subsection{Training Settings}
	We implement our model based on the Transformer library\footnote{\url{https://github.com/huggingface/transformers}}.
	We use the Adam \cite{kingma2014adam} optimizer to train models on each dataset for 5 epochs. The learning rate is set as 2$e$-5 and the dropout rate is set as 0.1. We use the largest batch size that can fit in the memory of GPU. 
	% The coefficient of L2 regularization is 0.01. 
	$\alpha$, $\beta$ and $\gamma$ are set as 0.9, 1, 0.01 respectively by our empirical experiments. % We use max pooling in the IC-Gated module.
	%and $K$ is set as 30\% of the length of all intents for important intents selection.
	All the experiments are conducted on V100 32GB GPUs.
	Each dataset is randomly split into training/validating/testing sets in the proportion of 8:1:1. All the parameters are tuned on the validation set and the results follow the 5-fold cross validation in testing set.
	
	% The root mean squad error (RMSE) is employed for NPS regression task to measure the error, and the smaller RMSE is the better.
	% Considering that our problem is from real-world and the data distribution is unbalanced, we use micro F1 score (Micro-F1) for category prediction task. The larger Micro-F1 is the better.
	
	\subsection{Baselines}
	We compare our proposed model to several baselines which belong to two groups. One includes single-task models for NPS prediction and dialog category classification separately, %like fastText\cite{Joulin2016}, HAN \cite{Yang2016} (we pad each utterance in a dialog as a long document for classification and regression respectively), textReg \cite{Dereli2019}, CNN-BiRNN-Att \cite{Yu2020}, and BERT \cite{Devlin2018}.
	while the other contains methods under multi-task learning paradigm. 
	To make a fair comparison between baselines and the G3M, we concatenate the role and intent information ahead each utterance only if the baseline model can employ that information. Thus all the baseline models have the consistent input information as ours.
	%In order to make sure the comparison of baselines and G3M model using the same data source, for all the baselines (apart from Dialog Router model), first we pad the role and intent information ahead each utterance in a dialog, then we pad the above utterance as a long document for classification and regression respectively. Note that all the baselines do not contain the gated mechanism.
	The evaluated baselines are listed as follows:
	\begin{itemize}
		\setlength{\itemsep}{0pt}
		\setlength{\parsep}{0pt}
		\setlength{\parskip}{0pt}
		\setlength{\itemindent}{0em}
		\item fastText~\cite{Joulin2016} is a simple and efficient model for regression or classification.
		%\item textCNN\cite{Kim2014,Zhang2015}, a convolution neural networks based model
		%\item textRNN\cite{Wang2018}, a bi-directional LSTM based model
		%\item HBLSTM \cite{Kumar2018}, is a hierarchical recurrent neural network based model, we modify the last layer to adapt both classification and regression tasks respectively.
		\item HAN \cite{Yang2016} is a document classification model based on hierarchical attention network by considering both word-level and sentence-level information.
		\item textReg \cite{Dereli2019} is a regression model based on convolution neural networks, which achieves the state-of-the-art performance on financial prediction.
		\item CNN-BiRNN-Att (CBA) \cite{Yu2020} is a cascade model with CNN, bi-directional RNN and attention mechanism. We adapt the last layer to support either the regression or classification tasks.
		\item BERT \cite{Devlin2018} is a model which can be fine-tuned by adjusting the task-specific output layers. We use the pre-trained BERT-base-cased model to separately fine-tune the two tasks with default parameters.
		\item XLNET~\cite{yang2019xlnet} is a generalized auto-regressive pre-training method which can be fine-tuned by tailing a task-specific output layers. We fine-tune the two tasks with this pre-trained model similar to BERT.
		\item Joint-CNN-BiRNN-Att (Joint-CBA) is a multi-task model by extending the last layer of CBA model to simultaneously predict NPS and dialog categories.
		\item Vanilla Multi-task Model (VMM) follows the vanilla multi-task learning paradigm and uses the standard BERT encoder to encode only the utterance data~\cite{huang2021dialog}, without any gated-mechanism modules.
		% ~\cite{huang2021dialog}, is the first work to try to learn a multi-task learning for dialog routing scenario with BERT as the encoder to learn dialog category and NPS tasks at the same time. We keep the same data preprocessing in this work to reproduce it and we pad intent information ahead each utterance in a dialog before goes into encoder. We keep the same result from this work in Learning dataset.
	\end{itemize}
	
	To measure the performance, we use the root mean squad error (RMSE) for the NPS regression task and micro F1 score (Micro-F1) for the dialog category classification task.

	\subsection{Results and Analysis}
	%\subsection{Main Performance}
	
	We compare our model with various single-task and multi-task baselines on the two datasets, and the first two sections of Table~\ref{table:comparison_of_learning} report the results. The performance of NPS regression task is reflected by RMSE metric, while that of dialog category classification task is embodied by Micro-F1 metric. We can learn some observations from Table~\ref{table:comparison_of_learning}. 
	
	Firstly, the performance of our proposed model G3M is consistent on both datasets and can outperform all the baselines. More specifically, for the NPS regression task, G3M's RMSE is 8.72\% and 11.83\% lower than the best RMSE score of baseline models (VMM for Learning dataset and Joint-CBA for MacHelp dataset) on two datasets. And  for the dialog category classification task, G3M upgrades the performance by achiving 2.17\% and 4.40\% higher Micro-F1 scores compared with the best baseline method (VMM). All the results demonstrate that our model is effective and achieves the state-of-the-art performance on both subtasks.
	
	Secondly, comparing the single-task models with multi-task models among baselines, it is consistent that multi-task models can surpass the single-task counterparts, especially from the comparison between Joint-CBA and VMM with their single-task versions (CBA and BERT). The results suggest that the underlying relation between the two tasks is helpful and multi-task learning paradigm can well capture the knowledge. 
	
	Thirdly, comparing three multi-task models, we observe that G3M is better than the other two methods. We can conclude that both our proposed dialog encoder and gated mechanism are effective, since Joint-CBA and VMM do not contain either of the modules. Another observation is that under the same multi-task learning paradigm, BERT based dialog encoder (VMM) is better than the CNN-RNN based dialog encoder (Joint-CBA).

		\begin{table}[]\scalebox{0.65}{
			\centering
			\begin{tabular}{|l|l|c|c|c|c|}
				\hline
				\multicolumn{2}{|c|}{\multirow{2}{*}{Models}} & \multicolumn{2}{c|}{Learning}    & \multicolumn{2}{c|}{MacHelp}     \\ \cline{3-6}
				\multicolumn{2}{|c|}{}                        &  RMSE       & Micro-F1           &  RMSE      &     Micro-F1        \\ \hline
				\multirow{6}{*}{Single-task} & fastText   & 0.9711    & 77.85                   & 1.2365    & 85.88      \\
				&HAN                                                  & 0.9981   & 78.85                  & 1.1276    & 84.19        \\
				&textReg                                             & 0.9689    & 77.98                    & 1.1099      & 85.55    \\
				&CBA                                              & 0.9697  & 79.42                & 1.1383    & 86.02         \\
				&BERT                                                  & 0.9702    & 80.04                 & 1.1025       & 86.39    \\
				&XLNET                                                  & 0.9648    & 80.21                 & 1.1084       & 86.25    \\\hline
				\multirow{3}{*}{Multi-task} &Joint-CBA                                   & 0.9555     & 80.38                    & 1.0624     & 87.21    \\
				&VMM                                           & 0.9302     & 81.88                 & 1.0767    & 87.49       \\
				&\textbf{G3M}                                         & \textbf{0.8491}  & \textbf{83.66}   & \textbf{0.9367} & \textbf{91.34} \\ \hline
				\multirow{3}{*}{Ablation} & - IC-Gated module                                       & 0.8892     & 82.57              & 0.9883    & 90.63       \\
				&- NC-Gated module                                       & 0.8878   & 82.73                & 0.9831    & 90.36       \\
				& - both gated modules                                      & 0.9187     & 82.01                & 1.0591   & 88.27        \\
				\hline
			\end{tabular}%
		}
		\caption{Comparison between G3M and baselines on Learning and MacHelp datasets. RMSE is for NPS prediction and Micro-F1 (\%) is for dialog classification.}
		\label{table:comparison_of_learning}
		% \vspace{-2ex}
	\end{table}
	\begin{figure}
		\begin{centering}
			\subfloat[Learning Corpus]{\begin{centering}
					\includegraphics[width=0.25\textwidth]{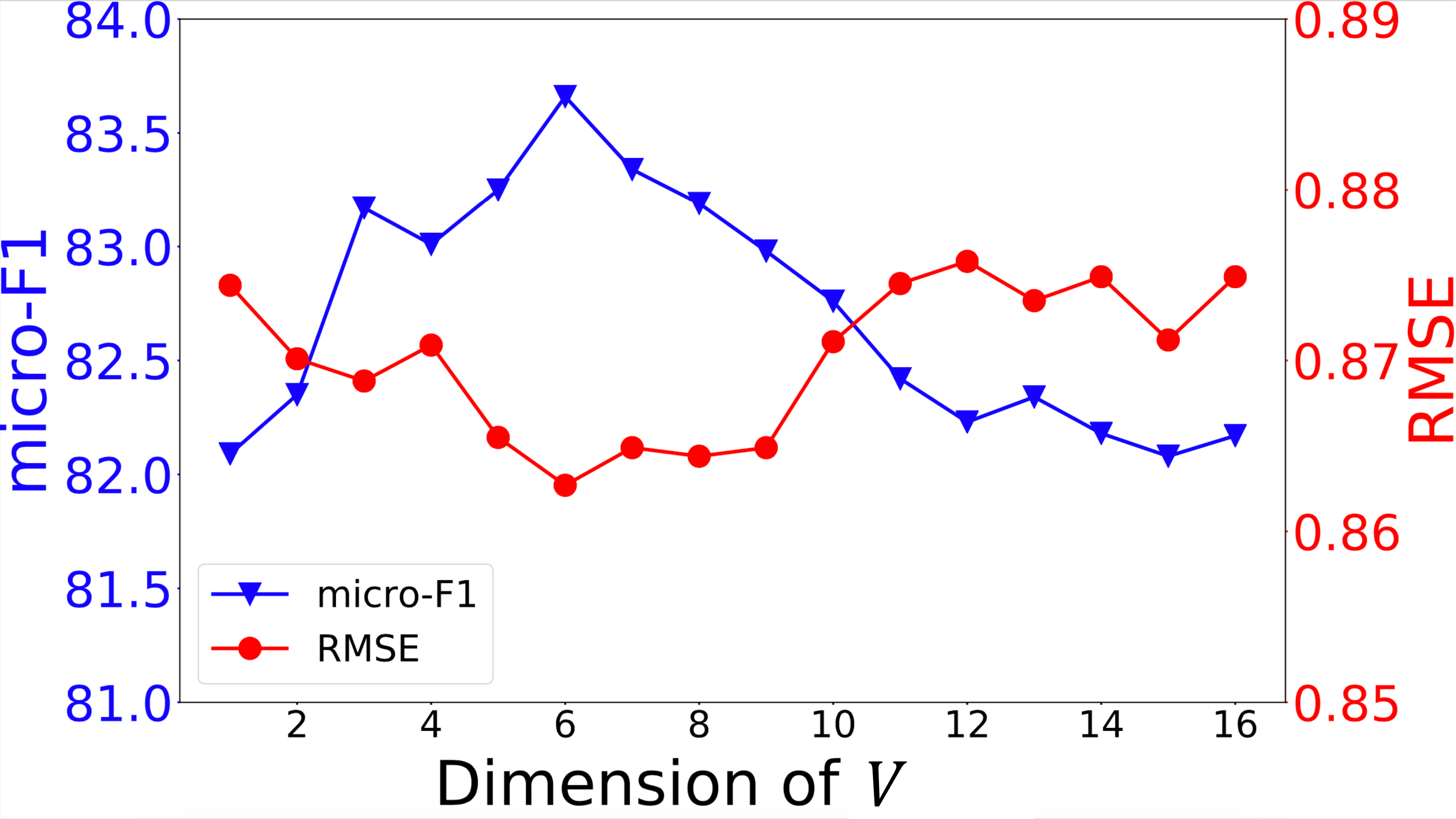}
					\par\end{centering}}
			\subfloat[MacHelp Corpus]{\begin{centering}
					\includegraphics[width=0.25\textwidth]{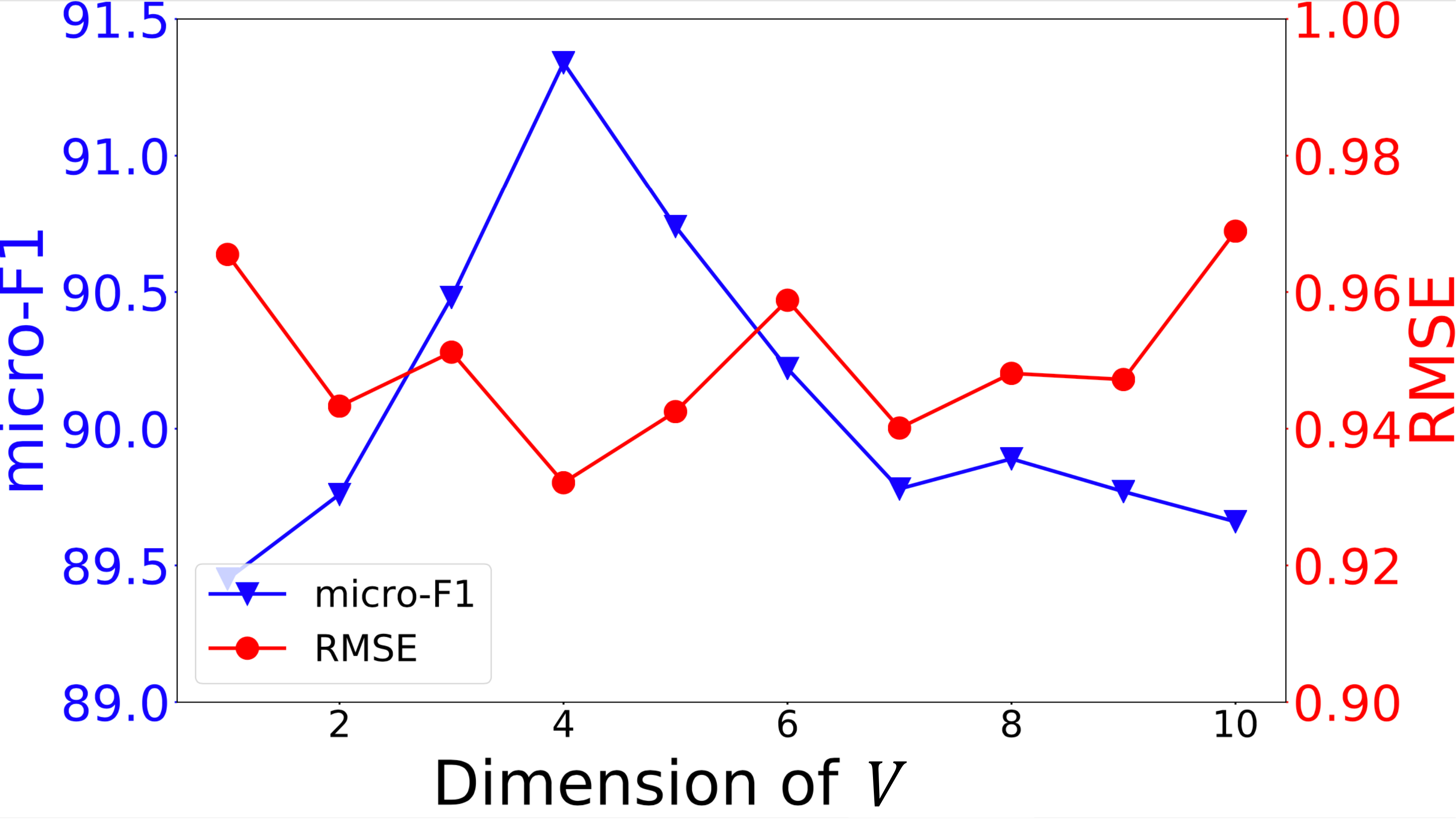}
					\par\end{centering}}    \par
		\end{centering}
		\caption{The parameter effect on different dimensions of $V$ in IC-Gated module of G3M. \label{fig:different_g}}
		\vspace{-3ex}
	\end{figure}
	
	\iffalse
	% \usepackage{multirow}
	\begin{table}[]\scalebox{0.9}{
			\centering
			\begin{tabular}{|l|c|c|c|c|}
				\hline
				\multicolumn{1}{|c|}{\multirow{2}{*}{Models}} & \multicolumn{2}{c|}{Learning}    & \multicolumn{2}{c|}{MacHelp}     \\ \cline{2-5}
				\multicolumn{1}{|c|}{}                        & Micro-F1       & RMSE            & Micro-F1       & RMSE            \\ \hline
				Non-Gated Model                               & 81.67          & 0.9312          & 88.27          & 1.0291          \\
				Non-IC-gate Model                             & 82.02          & 0.8892          & 89.36          & 0.9831          \\
				Non-NC-gate Model                             & 82.15          & 0.8978          & 89.63          & 0.9883          \\
				Ours                                          & \textbf{82.91} & \textbf{0.8627} & \textbf{90.74} & \textbf{0.9321} \\ \hline
		\end{tabular}}
		\caption{Ablation experiments by removing component from our proposed model}
		\label{table:ablation}
	\end{table}
	\fi
	
	\subsection{Ablation Study}
	
	To probe G3M's effectiveness in terms of gated module level, we conduct ablation experiments and report the results in the bottom section of Table~\ref{table:comparison_of_learning}. We could obtain three variant models by once deducting one or two gated modules: G3M without IC-Gated module, G3M without NC-Gated module, and G3M without either module. Based on the experiments, we have some findings.
	
	Comparing the three ablated variant models of G3M, both modules contribute positively to the two tasks. Therefore, they are effective to model the data-to-task and task-to-task information among dialog data. Specifically, it is not very clear that which module is more important than the other. However, ablating both them would damage the final performance.
	
	Furthermore, we compare the performance between G3M without both gated modules and multi-task baselines (Joint-CBA and VMM), we find the former model performs better than the later, which may indicate that our proposed dialog encoder is superior than the CNN-RNN based and BERT based dialog encoders.

	\subsection{Parameter Effect}
	\label{diff_dim} 
	
	In the IC-Gated module, there is an important parameter to be preset by humans, which is the dimension of $V$. It determines the size of weighted feature $g^{(I)}$ and therefore controls to what degree the data-to-task information can be leveraged. We investigate the effect of various dimensions of $V$ on both datasets. From the curves of RMSE and Micro-F1 scores in Figure~\ref{fig:different_g}, we find that the final performance is related to this parameter. Empirically, the optimal setting is 6 for Learning dataset and 4 for MacHelp dataset. We can also observe too small or too large dimensions may both play a harmful effect on extracting important knowledge from the data-to-task information. Another interesting phenomenon is that the optimal parameter values are coincidentally close or equal to the numbers of dialog categories to be classified, and we would explore the potential correlation in the future.

	\subsection{Case Study}
	% We do a case study to demonstrate our method's effectiveness on a complete conversation session between a user (U) and two human agents (A1 and A2) in \autoref{table:user_cases}. At first, human agent 1 serves the user and the NPS is from 5.8 to 5.0, which means the user is not satisfied with the agent. After the agent says she is not able to solve the problem, the number dropped to 5.0.  The conversation, therefore, is transferred to a more skilled human agent 2 and the NPS increases to 7.8 eventually. The human language between the user and agents can support that our model is reasonable to make every NPS prediction.
	We conduct a case study to demonstrate how our method could work to improve the system efficiency and user experience by integrating our dialog routing component into a human-bot symbiosis dialog system. Table~\ref{table:user_cases} lists the dialog utterances along with the predicted NPS and dialog category. Here the user (U) has a requirement to solve course enrollment issue. At first, the bot agent (A1) serves the user which is skilled at `Finding Content' dialogs, but it wrongly understands the user that she would find some course materials. By using the dialog routing component to monitor the dialogs, the predicted NPS decreases from 5.8 to 4.8, which means the user is dissatisfactory with the A1 and the routing to other agent should be triggered\footnote{We can set a dissatisfaction threshold based on real situations, e.g. 5 in this case.}. As a result, based on the dialog category classification by dialog routing component, the dialog is transferred to a human agent (A2) who is skilled at `Completion Issue' dialogs. Thanks to the right time to transfer and the correctly assigned agent, the NPS is up to 7.8 and the user is finally satisfactory with this service.
	
	%We do a case study to demonstrate our method's effectiveness on a complete conversation session between a user (U) and two agents (A1 means bot agent labelled with `Finding Content' and A2 means human agent labelled with `Completion Issue') 
	% (bot agents A1 and human agents A2) 
	%in \autoref{table:user_cases}. At first, bot agent A1 serves the user and the NPS is from 5.8 to 4.8, which means the user is not satisfied with this agent. After the agent says she is not able to solve the problem, the number dropped to 4.8.  The conversation, therefore, is transferred to a more skilled human agent A2 (the predicted current dialog category is `Completion Issue' and human agent A2 skilled at this category) and the NPS increases to 7.8 eventually. The human language between the user and agents can support that our model is reasonable to make every NPS prediction.
	
	This case could also reveal that our method would not affect the existing dialog systems and the dialog routing system can be deployed in a plug-in manner, which has a good compatibility and is convenient for both dialog routing system and dialog system to upgrade their ability.
	
	\begin{table}[t]\scalebox{0.75}{
			\begin{tabular}{|c|l|}
				\hline
				Role& Utterances \\
				\hline
				U & \tabincell{l}{Hi}                                                                          \\
				A1 & \tabincell{l}{Hello, how may I assist you?}                                                \\
				U & \tabincell{l}{I have enrolled in Deep Learning Course but today \\I can't see this enrollment} \\
				U & \tabincell{l}{I have completed more than 80\%}                                              \\
				U & \tabincell{l}{Can you please help me to find / resume the course}                            \\
				A1 & \tabincell{l}{I'm sorry that we are not able to do that}                                    \\
				U & \tabincell{l}{Why so}                                                                          \\
				U & \tabincell{l}{and what help you can do?}                                                \\\hline
				\multicolumn{2}{|l|}{Current NPS: 5.8}                                        \\
				\multicolumn{2}{|l|}{Current dialog category: Finding Content}                                        \\ \hline
				A1 & \tabincell{l}{You could get more information in the completion \\dashboard in the system} \\
				U & \tabincell{l}{You did not answer my question, I can't find the \\course and I want to resume it}                                              \\
				U & \tabincell{l}{Can you please help me to find / resume the course}                            \\
				A1 & \tabincell{l}{I'm sorry that we are not able to do that}                                    \\ \hline
				\multicolumn{2}{|l|}{Current NPS: 4.8 (\textbf{under the threshold and trigger routing})}                                         \\ 
				\multicolumn{2}{|l|}{Current dialog category: Completion Issue}                                        \\ \hline
				A2 & \tabincell{l}{Sorry for not being very helpful.}                                                                          \\
				A2 & \tabincell{l}{You can't find the course in your learning queue and\\ the progress are all gone, correct?
				}                                                \\
				U & \tabincell{l}{Exactly} \\
				\hline
				\multicolumn{2}{|l|}{Current NPS: 6.7}                                         \\ 
				\multicolumn{2}{|l|}{Current dialog category: Completion Issue}                                        \\ \hline
				A2 & \tabincell{l}{Is it an internal course in the Learning system, right?}                                              \\
				U & \tabincell{l}{Yes}                            \\
				\hline
				\multicolumn{2}{|l|}{Current NPS: 6.9}                                         \\ 
				\multicolumn{2}{|l|}{Current dialog category: Completion Issue}                                        \\ \hline
				A2 & \tabincell{l}{Sorry for your confusion, sometimes it may be \\caused  by technical issues.
				}                                              \\
				A2 & \tabincell{l}{A ticket has been created for you so that it will be \\reflected correctly.
				}                            \\
				A2 & \tabincell{l}{Normally the technical team will get back to you \\in 24 hours
				}                            \\
				U & \tabincell{l}{Thanks, bye
				}                            \\
				A2 & \tabincell{l}{Bye
				}                            \\
				\hline
				\multicolumn{2}{|l|}{Current NPS: 7.8}                                        \\ 
				\multicolumn{2}{|l|}{Current dialog category: Completion Issue}                                        \\ \hline
		\end{tabular}}
		\caption{A dialog session along with the predicted NPS and dialog category by equipping a dialog routing component with our method.} %U means user, A1 means bot agent labelled with Finding\_Content and A2 means human agent 2 labelled with Completion\_Issue. }
		%We empirically set NPS $<=$ 5 represents dissatisfaction.}
		\label{table:user_cases}
		\vspace{-3ex}
	\end{table}

	\section{Conclusion}
	
	In current ubiquitous human-bot symbiosis dialog systems for customer care, the dialog routing component is necessary to improve the overall system efficiency, reduce human resource cost, and enhance user experience. In this paper, we argue that the data-to-task and task-to-task information among various kinds of dialog data and subtasks should be jointly leveraged to perform better dialog routing ability. We propose a Gated Mechanism enhanced Multi-task Model (G3M) to implement that motivation. Specifically, we design a new dialog encoder to learn various kinds of dialog data by extending the BERT encoder, and two gated mechanism modules are proposed to capture data-to-task and task-to-task information. Extensive experiments on two real-world datasets demonstrate the effectiveness of our proposed model, which can achieve the state-of-the-art performance.
	
	%In human-bot symbiosis dialog systems, the reliable dialog routing component is necessary to improve the overall system’s efficiency and user's satisfaction. In this paper, we investigate the less-studied problem that how to route the conversation among bot and human agents in a more precise way. We propose a Gated Mechanism enhanced Multi-task Model (G3M) on BERT to fully leverage the underlying information between task to task and data to task. Extensive experiments on two real-world datasets demonstrate the proposed model can achieve the state-of-the-art accuracy. With the reliable dialog NPS prediction and category classification, the rules to trigger routing at every dialog turn can be more easily set by the actual dialog scenario to facilitate the most existing human-bot hybrid dialog systems for customer care.

	\section*{Acknowledgement}
    We thank all the anonymous reviewers for their valuable feedback and insightful comments. This work is partially supported by the Natural Science Foundation of Jiangsu Province, China (Grant No. BK20220488). 
	
	\iffalse
	\section*{Acknowledgements}
	
	This document has been adapted
	by Steven Bethard, Ryan Cotterell and Rui Yan
	from the instructions for earlier ACL and NAACL proceedings, including those for 
	ACL 2019 by Douwe Kiela and Ivan Vuli\'{c},
	NAACL 2019 by Stephanie Lukin and Alla Roskovskaya, 
	ACL 2018 by Shay Cohen, Kevin Gimpel, and Wei Lu, 
	NAACL 2018 by Margaret Mitchell and Stephanie Lukin,
	Bib\TeX{} suggestions for (NA)ACL 2017/2018 from Jason Eisner,
	ACL 2017 by Dan Gildea and Min-Yen Kan, 
	NAACL 2017 by Margaret Mitchell, 
	ACL 2012 by Maggie Li and Michael White, 
	ACL 2010 by Jing-Shin Chang and Philipp Koehn, 
	ACL 2008 by Johanna D. Moore, Simone Teufel, James Allan, and Sadaoki Furui, 
	ACL 2005 by Hwee Tou Ng and Kemal Oflazer, 
	ACL 2002 by Eugene Charniak and Dekang Lin, 
	and earlier ACL and EACL formats written by several people, including
	John Chen, Henry S. Thompson and Donald Walker.
	Additional elements were taken from the formatting instructions of the \emph{International Joint Conference on Artificial Intelligence} and the \emph{Conference on Computer Vision and Pattern Recognition}.
	\fi
	
	\bibliography{coling22}

\begin{thebibliography}{20}
\expandafter\ifx\csname natexlab\endcsname\relax\def\natexlab#1{#1}\fi

\bibitem[{Chen et~al.(2018)Chen, Yang, Liu, Zhao, and Jia}]{Yubo2018}
Yubo Chen, Hang Yang, Kang Liu, Jun Zhao, and Yantao Jia. 2018.
\newblock Collective event detection via a hierarchical and bias tagging
  networks with gated multi-level attention mechanisms.
\newblock In \emph{Proceedings of the 2018 Conference on Empirical Methods in
  Natural Language Processing}, pages 1267--1276.

\bibitem[{Cohan et~al.(2019)Cohan, Beltagy, King, Dalvi, and
  Weld}]{cohan2019pretrained}
Arman Cohan, Iz~Beltagy, Daniel King, Bhavana Dalvi, and Daniel~S Weld. 2019.
\newblock Pretrained language models for sequential sentence classification.
\newblock \emph{arXiv preprint arXiv:1909.04054}.

\bibitem[{Dereli and Sara{\c{c}}lar(2019)}]{Dereli2019}
Ne{\c{s}}at Dereli and Murat Sara{\c{c}}lar. 2019.
\newblock Convolutional neural networks for financial text regression.
\newblock In \emph{Proceedings of the 57th Annual Meeting of the Association
  for Computational Linguistics: Student Research Workshop}, pages 331--337.

\bibitem[{Devlin et~al.(2019)Devlin, Chang, Lee, and Toutanova}]{Devlin2018}
Jacob Devlin, Ming-Wei Chang, Kenton Lee, and Kristina Toutanova. 2019.
\newblock \href {https://doi.org/10.18653/v1/N19-1423} {{BERT}: Pre-training of
  deep bidirectional transformers for language understanding}.
\newblock In \emph{Proceedings of the 2019 Conference of the North {A}merican
  Chapter of the Association for Computational Linguistics: Human Language
  Technologies, Volume 1 (Long and Short Papers)}, pages 4171--4186,
  Minneapolis, Minnesota. Association for Computational Linguistics.

\bibitem[{Du et~al.(2019)Du, Huang, He, Liu, and Zhu}]{du2019knowledge}
Zefeng Du, Peijie Huang, Yuhong He, Wei Liu, and Jiankai Zhu. 2019.
\newblock A knowledge-gated mechanism for utterance domain classification.
\newblock In \emph{CCF International Conference on Natural Language Processing
  and Chinese Computing}, pages 142--154. Springer.

\bibitem[{Goo et~al.(2018)Goo, Gao, Hsu, Huo, Chen, Hsu, and
  Chen}]{goo2018slot}
Chih-Wen Goo, Guang Gao, Yun-Kai Hsu, Chih-Li Huo, Tsung-Chieh Chen, Keng-Wei
  Hsu, and Yun-Nung Chen. 2018.
\newblock Slot-gated modeling for joint slot filling and intent prediction.
\newblock In \emph{Proceedings of the 2018 Conference of the North American
  Chapter of the Association for Computational Linguistics: Human Language
  Technologies, Volume 2 (Short Papers)}, pages 753--757.

\bibitem[{Huang et~al.(2021)Huang, Jiang, Chen, Han, and
  Dang}]{huang2021dialog}
Ziming Huang, Zhuoxuan Jiang, Hao Chen, Xue Han, and Yabin Dang. 2021.
\newblock Dialog router: Automated dialog transition via multi-task learning.
\newblock In \emph{Proceedings of the AAAI Conference on Artificial
  Intelligence}, volume~35, pages 16038--16040.

\bibitem[{Ide and Kawahara(2021)}]{naacl21}
Tatsuya Ide and Daisuke Kawahara. 2021.
\newblock \href {http://arxiv.org/abs/2105.11696} {Multi-task learning of
  generation and classification for emotion-aware dialogue response
  generation}.
\newblock \emph{CoRR}, abs/2105.11696.

\bibitem[{Joulin et~al.(2017)Joulin, Grave, Bojanowski, Douze, J{\'e}gou, and
  Mikolov}]{Joulin2016}
Armand Joulin, Edouard Grave, Piotr Bojanowski, Matthijs Douze, H{\'e}rve
  J{\'e}gou, and Tomas Mikolov. 2017.
\newblock Fasttext. zip: Compressing text classification models.
\newblock In \emph{The International Conference on Learning Representations
  (ICLR)}.

\bibitem[{Kim(2014)}]{Kim2014}
Yoon Kim. 2014.
\newblock Convolutional neural networks for sentence classification.
\newblock In \emph{Conference on Empirical Methods in Natural Language
  Processing}.

\bibitem[{Kingma and Ba(2014)}]{kingma2014adam}
Diederik~P Kingma and Jimmy Ba. 2014.
\newblock Adam: A method for stochastic optimization.
\newblock \emph{arXiv preprint arXiv:1412.6980}.

\bibitem[{Kumar et~al.(2018)Kumar, Agarwal, Dasgupta, and Joshi}]{Kumar2018}
Harshit Kumar, Arvind Agarwal, Riddhiman Dasgupta, and Sachindra Joshi. 2018.
\newblock Dialogue act sequence labeling using hierarchical encoder with crf.
\newblock In \emph{Thirty-Second AAAI Conference on Artificial Intelligence}.

\bibitem[{Ngo-Ye and Sinha(2014)}]{ngo2014influence}
Thomas~L Ngo-Ye and Atish~P Sinha. 2014.
\newblock The influence of reviewer engagement characteristics on online review
  helpfulness: A text regression model.
\newblock \emph{Decision Support Systems}, 61:47--58.

\bibitem[{Oraby et~al.(2017)Oraby, Gundecha, Mahmud, Bhuiyan, and
  Akkiraju}]{oraby2017may}
Shereen Oraby, Pritam Gundecha, Jalal Mahmud, Mansurul Bhuiyan, and Rama
  Akkiraju. 2017.
\newblock How may i help you?: Modeling twitter customer serviceconversations
  using fine-grained dialogue acts.
\newblock In \emph{Proceedings of the 22nd International Conference on
  Intelligent User Interfaces}, pages 343--355. ACM.

\bibitem[{Singh et~al.(2015)Singh, Verma, and Thoke}]{Singh2015}
Bikesh~Kumar Singh, Kesari Verma, and AS~Thoke. 2015.
\newblock Investigations on impact of feature normalization techniques on
  classifier's performance in breast tumor classification.
\newblock \emph{International Journal of Computer Applications}, 116(19).

\bibitem[{Wang et~al.(2018)Wang, Liu, Luo, and Wang}]{Wang2018}
Jenq-Haur Wang, Ting-Wei Liu, Xiong Luo, and Long Wang. 2018.
\newblock An lstm approach to short text sentiment classification with word
  embeddings.
\newblock In \emph{Proceedings of the 30th conference on computational
  linguistics and speech processing (ROCLING 2018)}, pages 214--223.

\bibitem[{Xiao et~al.(2018)Xiao, Zhang, and Chen}]{Xiao18}
Liqiang Xiao, Honglun Zhang, and Wenqing Chen. 2018.
\newblock \href {https://doi.org/10.18653/v1/n18-2114} {Gated multi-task
  network for text classification}.
\newblock In \emph{Proceedings of the 2018 Conference of the North American
  Chapter of the Association for Computational Linguistics: Human Language
  Technologies, NAACL-HLT, New Orleans, Louisiana, USA, June 1-6, 2018, Volume
  2 (Short Papers)}, pages 726--731. Association for Computational Linguistics.

\bibitem[{Yang et~al.(2019)Yang, Dai, Yang, Carbonell, Salakhutdinov, and
  Le}]{yang2019xlnet}
Zhilin Yang, Zihang Dai, Yiming Yang, Jaime Carbonell, Russ~R Salakhutdinov,
  and Quoc~V Le. 2019.
\newblock Xlnet: Generalized autoregressive pretraining for language
  understanding.
\newblock In \emph{Advances in neural information processing systems}, pages
  5753--5763.

\bibitem[{Yang et~al.(2016)Yang, Yang, Dyer, He, Smola, and Hovy}]{Yang2016}
Zichao Yang, Diyi Yang, Chris Dyer, Xiaodong He, Alex Smola, and Eduard Hovy.
  2016.
\newblock Hierarchical attention networks for document classification.
\newblock In \emph{Proceedings of the 2016 conference of the North American
  chapter of the association for computational linguistics: human language
  technologies}, pages 1480--1489.

\bibitem[{Yu et~al.(2020)Yu, Guan, Ma, Jiang, and Huang}]{Yu2020}
Yipeng Yu, Ran Guan, Jie Ma, Zhuoxuan Jiang, and Jingchang Huang. 2020.
\newblock \href {https://www.aclweb.org/anthology/2020.coling-main.358/} {When
  and who? conversation transition based on bot-agent symbiosis learning
  network}.
\newblock In \emph{Proceedings of the 28th International Conference on
  Computational Linguistics, {COLING} 2020, Barcelona, Spain (Online), December
  8-13, 2020}, pages 4056--4066. International Committee on Computational
  Linguistics.

\end{thebibliography}
	\bibliographystyle{acl_natbib}
	
	\iffalse
	\appendix
	
	\section{Example Appendix}
	\label{sec:appendix}
	
	This is an appendix.
	\fi
	
\end{document}